\documentclass[10pt,twocolumn,letterpaper]{article}

\usepackage{iccv}
\usepackage{times}
\usepackage{epsfig}
\usepackage{graphicx}
\usepackage{amsmath}
\usepackage{amssymb}

\usepackage[pagebackref=true,breaklinks=true,letterpaper=true,colorlinks,bookmarks=false]{hyperref}

\iccvfinalcopy 

\begin{document}

\title{Leveraging Image-Text Similarity and Caption Modification for the DataComp Challenge: Filtering Track and BYOD Track}

\author{Shuhei Yokoo\thanks{These authors contributed equally to this work.}, Peifei Zhu\footnotemark[1], Yuchi Ishikawa, Mikihiro Tanaka, Masayoshi Kondo, Hirokatsu Kataoka \\
LINE Corporation\thanks{Now known as LY Corporation since October 2023.} \\
{\tt\small \{shuhei.yokoo, peifei.zhu, yuchi.ishikawa, mikihiro.tanaka, masayoshi.kondo, jpz4219\}} \\ {\tt\small@linecorp.com}
}
\maketitle

\begin{abstract}
Large web crawl datasets have already played an important role in learning multimodal features with high generalization capabilities. However,  there are still very limited studies investigating the details or improvements of data design. Recently, a DataComp challenge has been designed to propose the best training data with the fixed models. This paper presents our solution to both filtering track and BYOD track of the DataComp challenge. Our solution adopts large multimodal models CLIP and BLIP-2 to filter and modify web crawl data, and utilize external datasets along with a bag of tricks to improve the data quality. Experiments show our solution significantly outperforms DataComp baselines (filtering track: 6.6\% improvement, BYOD track: 48.5\% improvement).
\end{abstract}

\section{Introduction}

In recent years, multimodal models such as CLIP \cite{radford2021learning}, BLIP-2 \cite{li2023blip}, and Stable Diffusion \cite{rombach2022high} have been proposed and achieved state-of-the-art generalization capabilities in downstream tasks such as zero-shot classification and text-guided image generation. A common feature of these models is they were trained on hundreds of millions of web crawl data.
For example, CLIP was trained on 400 million image-text pairs collected from the internet, and Stable Diffusion was trained on LAION-5B~\cite{laion5b} which contains 5.8 billion image-text pairs.

Even though large datasets have made breakthroughs in multimodal learning, there are still very limited studies exploring how data choices would affect the final performance. In particular, web crawl datasets usually contain numerous low-quality samples that can mislead and negatively affect the performance of the model. How to achieve satisfactory results by designing high-quality datasets has gained increasing attention.

The DataComp challenge \cite{gadre2023datacomp} at ICCV’23 was organized as a competition for proposing new training sets while fixing the models and training code. There were two tracks in this challenge: filtering track and BYOD track. In the filtering track, participants were asked to design new filtering techniques. Meanwhile, in the BYOD track, adding existing datasets was allowed to curate new data sources. 
Following the development in either track, participants then evaluate the new dataset by running a standardized CLIP training code. The final step was to test the model on 38 downstream test sets.

In this paper, we propose a bag of tricks to design new training sets for the CLIP model. Our approach achieved better results than DataComp baselines in both tracks. We summarized our solution and findings as follows:

\begin{enumerate}
    \item For the filtering track, we have calculated image-text similarity for several models, including CLIP, and evaluated their scores on filtering tracks.
    We found two insights: (1) models that perform well on downstream tasks are not necessarily suitable for filtering, whereas (2) fine-tuning on a clean dataset such as MSCOCO is effective for filtering. The best result was achieved using BLIP-2 fine-tuned on MSCOCO (BLIP-2-COCO). Our final solution is selecting the samples with top 35\% image-text similarity calculated by BLIP-2-COCO.
    \item For the BYOD track, we conduct a caption modification using BLIP-2 to improve the consistency of the image-text pairs. In addition, we prepare several external datasets including a public image-text dataset CC12M and several training sets of the evaluation sets used in the downstream tasks. We also combine tricks such as using caption templates and dataset upsampling to further adjust our data. Experiments show our solution can significantly outperform the DataComp baseline for both small and medium scales.
\end{enumerate}

\section{Solution: Filtering Track}
For the filtering track, selecting image and text pairs that are well aligned is crucial. In this study, we investigated two hypotheses: (1) Can we enhance filtering performance by employing model effective at measuring image-text similarity, i.e., models that perform well in downstream tasks (such as image-text retrieval and ImageNet zero-shot classification)? (2) Can we improve filtering performance by using models that learn the correct correspondence between an image and a text on a clean dataset such as MSCOCO?

Regarding (1), contrary to our expectations, we did not observe a correlation between performance of downstream tasks and filtering effectiveness.
Concerning (2), when using BLIP-2~\cite{li2023blip} as a baseline, fine-tuned version with the MSCOCO~\cite{chen2015microsoft}~(BLIP-2-COCO) demonstrated superior performance.
This suggests that clean data even on a small scale can be effectively utilized for filtering.


BLIP-2~\cite{li2023blip} is a recently introduced vision and language foundation model, which achieves state-of-the-art results in several downstream tasks.
This work utilizes a pretrained image model and a large language model off-the-shelf, and a module called ``Q-Former'' enables to align their modalities.
For the final solution in the filtering track, we selected the top 35\% of samples based on the image-text similarity calculated by BLIP-2-COCO. We also tested combining BLIP-2-COCO with other approaches which have some positive effects solely (e.g., ensemble of image-text similarities, de-duplication), but these did not lead to any improvements.

\section{Solution: BYOD Track}
Our solution uses both the common pool data and external datasets. For the common pool data, we conduct caption modification using BLIP-2, while for the external datasets, we apply CLIP score filtering and propose several tricks. The details are as follows.
\subsection{Caption Modification using BLIP-2}
As the image-text pairs in the common pool are collected from the web, many samples contain texts that cannot accurately describe the images. This inconsistency has a harmful inference on learning vision and language alignment. Inspired by the CapFilt method \cite{li2022blip, li2023blip}, we conduct a caption modification to improve the consistency of the image-text pairs. For each data in the common pool, we first generate a caption using BLIP-2 equipped Flan-T5-XL~\cite{FLANChung2022ScalingIL}. Then we compare the CLIP ViT-B/32 score of the BLIP-2 caption and the original caption, and we choose the one with the higher CLIP score as the final caption. 

However, even with caption modification, there are still samples with low-resolution, noisy images left. Therefore, we further apply filtering on the modified dataset to select an optimal subset. We experimented with filtering using CLIP ViT-B/32 score under different thresholds and observed choosing the top 50\% of the data obtains the best result.

\subsection{External Datasets}

\noindent{\bf{Public image-text datasets.}} We use CC12M \cite{changpinyo2021conceptual}, a dataset with 12 million image-text pairs. Since the total amount of training computes remains constant (e.g. 12.8M samples seen for small scale), we only select a subset of the CC12M samples considering the size of the filtered common pool. We also use CLIP ViT-B/32 score for filtering and choose the top 30\% data for small scale and the top 50\% data for medium scale.

\noindent{\bf{Training data of evaluation sets.}} To improve the performance of downstream tasks on the evaluation sets, we use the supervised training sets of the downstream datasets. We choose the datasets with their training and testing sets already split, and we only use the training set to train the model. We experimented adding 15 datasets, including CIFAR-10 \cite{krizhevsky2009learning}, CIFAR-100 \cite{krizhevsky2009learning}, Country211 \cite{radford2021learning, thomee2016yfcc100m}, Food-101 \cite{bossard2014food}, GTSRB~\cite{stallkamp2011german}, ImageNet-1K~\cite{deng2009imagenet}, MNIST~\cite{lecun1998mnist}, Oxford Flowers-102 \cite{nilsback2008automated}, Pascal VOC 2007~\cite{everingham2010pascal}, PatchCamelyon \cite{veeling2018rotation, zhai2019visual}, Rendered SST2 \cite{zhai2019visual}, STL-10 \cite{coates2011analysis}, SVHN \cite{netzer2011reading}, Flickr30k~\cite{young2014image}, MSCOCO \cite{chen2015microsoft}. Some datasets (e.g. CIFAR-10, MNIST, PatchCamelyon) have a positive effect on the model performance, while others do not have an obvious effect (e.g. Country211, Rendered SST2) or have a negative effect (e.g. SVHN). We consider increasing the variety of the training data will have a positive effect, e.g. histopathology images rarely appear in the web data so adding PatchCamelyon largely improves the performance of the corresponding downstream task. Ultimately, we use the training data of MNIST, CIFAR-10, CIFAR-100, STL-10, PatchCamelyon, GTSRB, Oxford Flowers-102, and Pascal VOC 2007.

\subsection{Other Tricks}

\noindent{\bf{Caption template.}} For the training data of the evaluation sets, most of them are classification datasets and only have the class name as their label. To improve the model learning the alignment of image and text, we designed several caption templates for the training datasets. We notice that various prompt templates were already designed for zero-shot evaluation. On the other hand, we observe that the order of the words in the caption seems to have a large influence on the CLIP score, especially when words closer to the beginning have a greater influence. Therefore, we made some adjustments on the zero-shot prompt templates to obtain the caption templates, e.g. for GTSRB, instead of using ``a photo of a \{c\} traffic sign'', we use``traffic sign: \{c\}''. A list of the datasets we used and their caption template is shown in Table \ref{table:caption_templates}.

\begin{table}[t]
    \centering
    \caption{A list of caption templates. \{c\} stands for the ground-truth class label.}
    \vspace{2mm}
    \label{table:caption_templates}
    \begin{tabular}{ll}
        \hline
        Dataset & Caption template \\
        \hline
        MNIST & a photo of the number: \{c\} \\
        CIFAR-10 & a photo of \{c\} \\
        CIFAR-100 & a photo of \{c\} \\
        STL-10 & a photo of \{c\} \\
        PatchCamelyon & a histopathology image of \{c\} \\
        GTSRB & a traffic sign: \{c\} \\
        Oxford Flowers-102 & a type of flower: \{c\} \\
        Pascal VOC 2007 & a photo of \{c\} \\
        \hline
    \end{tabular}
\end{table}

\noindent{\bf{Dataset upsampling.}} As a type of data augmentation,  upsampling a specific dataset could improve the performance of that dataset. However, upsampling a dataset too many times might have a negative impact on the performance of other test datasets. We experimented with upsampling MNIST with different factors ($\times 2$, $\times 3$, $\times 4$). We found upsampling 3 times can largely improve the result of MNIST (accuracy from 0.43 to 0.75), as well as positively affect the average scores (shown in Table \ref{table:BYOD_small_scale}).

\section{Experiments}
\subsection{DataComp Datasets}
DataComp provides a new candidate pool of 12.8B image-text pairs collected from Common Crawl. This challenge contains four distinct compute and data scales. In this work, we mainly focus on the small scale (12.8M samples). Additionally, medium scale (128M samples) is also tested for the BYOD track.

\subsection{Filtering Results}
We experimented using several models capable of calculating image-text similarities, and evaluated the accuracy of filtering based on the similarities.
In BLIP-2 and BLIP-2-COCO, since the image-text similarity is calculated for each query of the Q-former, we adopted the maximum value of them as its image-text similarity.
The results are shown in Table~\ref{table:vlm_comparison}.
Interestingly, we found that the downstream task performances of similarity calculation models do not proportionally correlate with the score of filtering track.
For instance, CLIP ViT-B/32~\cite{radford2021learning} is relatively lower performance in terms of the downstream task performance, but the average score of filtering track using CLIP ViT-B/32 is the second highest in Table~\ref{table:vlm_comparison}.
The only exception was BLIP-2-COCO, which is superior to other models in terms of the downstream task performance, achieved the highest average score in filtering track.

\begin{table}[t]
    \centering
    \caption{
    The results of the model comparison on filtering track small scale.
    For all the models, samples that have top 30\% image-text similarities are selected.
    Models marked with ``openai'' are trained by OpenAI, marked with ``open\_clip'' employed models trained by~\cite{ilharco_gabriel_2021_5143773}.
    }
    \vspace{2mm}
    \label{table:vlm_comparison}
    \begin{tabular}{lcc}
        \hline
        Model & IN-1K & Average \\
        \hline
        CLIP ViT-B/32 (openai)~\cite{radford2021learning} & 0.055 & 0.171 \\
        CLIP ViT-B/16 (openai)~\cite{radford2021learning} & 0.049 & 0.162 \\
        CLIP ViT-L/14 (openai)~\cite{radford2021learning} & 0.050 & 0.166 \\
        CLIP ViT-L/14-336 (openai)~\cite{radford2021learning} & 0.044 & 0.163 \\
        CLIP-ViT-bigG/14 (open\_clip)~\cite{ilharco_gabriel_2021_5143773} & 0.042 & 0.154 \\
        CoCa-ViT-L/14 (open\_clip)~\cite{ilharco_gabriel_2021_5143773,yu2022coca} & 0.043 & 0.154 \\
        ConvNeXt-L (open\_clip)~\cite{ilharco_gabriel_2021_5143773,liu2022convnext} & 0.044 & 0.160 \\
        ConvNeXt-XXL (open\_clip)~\cite{ilharco_gabriel_2021_5143773,liu2022convnext} & 0.041 & 0.155 \\
        Roberta-ViT-B/32 (open\_clip)~\cite{ilharco_gabriel_2021_5143773} & 0.047 & 0.161 \\
        ImageBind-H~\cite{girdhar2023imagebind} & 0.045 & 0.159 \\
        BLIP-2~\cite{li2023blip} & 0.047 & 0.164 \\
        BLIP-2-COCO~\cite{li2023blip} & \textbf{0.056} & \textbf{0.175} \\
        \hline
    \end{tabular}
\end{table}

\begin{table}[t]
    \centering
    \caption{The comparison between BLIP-2-COCO and the baseline on filtering track small scale.}
    \vspace{2mm}
    \label{table:BLIP-2_30_vs_35}
    \begin{tabular}{lcc}
        \hline
         & IN-1K & Average \\
        \hline
        CLIP ViT-L/14 30\% (from \cite{gadre2023datacomp}) & 0.051 & 0.173 \\
        CLIP ViT-L/14 30\% (our repro.) & 0.050 & 0.166 \\
        \hline
        BLIP-2-COCO 35\% (ours) & \textbf{0.053} & \textbf{0.177} \\
        \hline
    \end{tabular}
\end{table}

Table~\ref{table:BLIP-2_30_vs_35} is our final result in small scale which contains the comparison to the baselines.
We experimented with similarity filtering thresholds at 25, 30, 35, and 40\% and adopted the best threshold: 35\%.
We tried to reproduce the official baseline result (``Baseline: CLIP score (L/14 30\%)''), but obtained slightly lower accuracy. For reference, the official baseline scores (from \cite{gadre2023datacomp}) is included in Table~\ref{table:BLIP-2_30_vs_35}, besides our reproduction results (our repro.).
Our result shows a 6.6\% improvement over the baseline, demonstrating our solution is effective despite its simplicity.

\begin{figure}[t]
   \begin{center}
      \includegraphics[width=0.95\linewidth, height=0.95\linewidth]{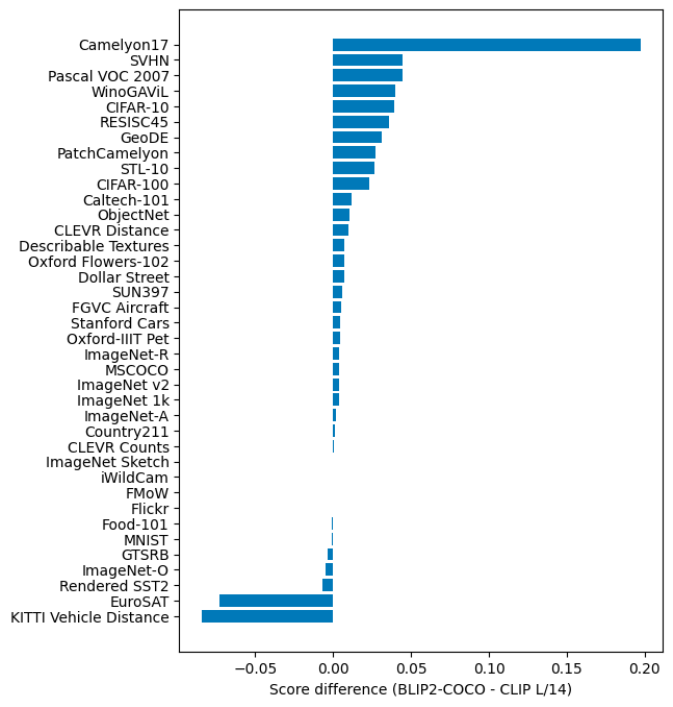}
   \end{center}
      \caption{Visualization of the score difference between the baseline (CLIP ViT-L/14 30\%) and our solution (BLIP-2-COCO 35\%) in the filtering track, small scale.}
   \label{fig:comparison_filtering}
\end{figure}

In Figure \ref{fig:comparison_filtering}, we visualize the comparison result between the baseline (CLIP ViT-L/14 30\%) and our final solution (BLIP-2-COCO 35\%) on 38 evaluation datasets. 

\subsection{BYOD Results}
\begin{figure}[t]
   \begin{center}
      \includegraphics[width=\linewidth, height=\linewidth]{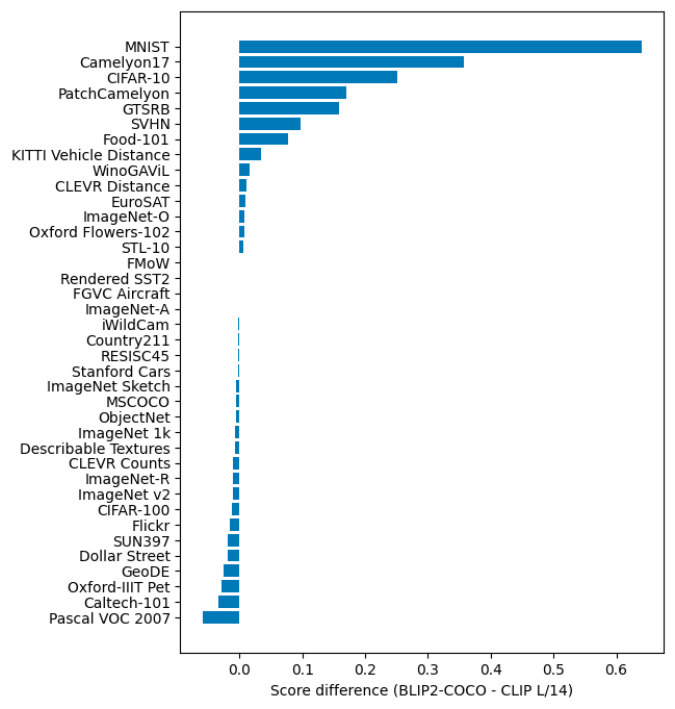}
   \end{center}
      \caption{Visualization of the score difference before and after adding the training data of evaluation sets in the BYOD track, small scale.}
   \label{fig:comparison_byod}
\end{figure}

\noindent{\bf{Small scale.}} We provide a step-wise comparison with the baseline which uses CLIP ViT-L/14 score to select 30\% of the common pool. The results are shown in Table \ref{table:BYOD_small_scale}. Caption modification and adding external datasets largely improved the performance of downstream tasks. We also provide a comparison of before and after adding the training data of evaluation sets (including MNIST upsampling) in Figure \ref{fig:comparison_byod}. The scores of the datasets such as MNIST, Camelyon17 were largely improved after adding their training sets. On the other hand, for Pascal VOC 2007 and CIFAR-100, although the scores of these datasets themselves slightly decreased, adding their training sets improved the average score. 

\begin{table*}[t]
    \centering
    \caption{The results of BYOD small scale. Ablations from the baseline to our final solution are provided.}
    \vspace{2mm}
    \label{table:BYOD_small_scale}
    \begin{tabular}{lccc}
        \hline
        & Data size & IN-1K & Average \\
        \hline
        baseline (CLIP ViT-L/14 30\%) & 3.8M & 0.047 & 0.163 \\
        BLIP caption modification (50\%) & 6.1M & 0.074 & 0.197 \\
        BLIP caption modification (50\%) + CC12M (30\%) & 8.9M & \textbf{0.080} & 0.201 \\
        BLIP caption modification (50\%) + CC12M (30\%) + Eval\_trainsets & 9.5M & 0.074 & 0.235 \\
        BLIP caption modification (50\%) + CC12M (30\%) + Eval\_trainsets (MNIST$\times$3) & 9.6M & 0.072 & \textbf{0.242} \\
        \hline
    \end{tabular}
\end{table*}

\begin{table*}[t]
    \centering
    \caption{The results of BYOD medium scale. CLIP score and image-based clustering are used for filtering the common pool, and the external datasets are the same with the small scale.}
    \vspace{2mm}
    \label{table:BYOD_medium_scale}
    \begin{tabular}{lccc}
        \hline
        & Data size & IN-1k & Average \\
        \hline
        baseline (CLIP ViT-L/14 30\%) & 36M & 0.269 & 0.324 \\
        CLIP  (30\%) + CC12M (50\%) + Eval\_trainsets (MNIST$\times$3) & 42M & 0.285 & 0.390 \\
        Image-cluster and CLIP (40\%) + CC12M (50\%) + Eval\_trainsets (MNIST$\times$3) & 22M & \textbf{0.326} & \textbf{0.398} \\
        \hline
    \end{tabular}
\end{table*}

\noindent{\bf{Medium scale.}} For the medium scale, we first select data from the common pool in two ways: 1) CLIP ViT-L/14 (top 30\%), 2) image-based clustering filtering and CLIP ViT-L/14 score top 40\%. Then we add the external datasets (same as small scale) to construct the training data. The results of BYOD medium scale are shown in Table \ref{table:BYOD_medium_scale}. By combining image-based clustering, we achieved better results even with only half of the data.

\section{Conclusion}

This paper presents our solution to both filtering and BYOD tracks of the DataComp challenge. We proposed several simple yet powerful tricks to improve the training set. For the filtering track, we employ BLIP-2-COCO to calculate image-text similarity and select samples based on this similarity. For the BYOD track, we modify the caption in the common pool and make use of the external datasets with extra tracks such as adding caption templates and upsampling. Experiments show our solution significantly outperforms the baseline in both tracks.

{\small
\bibliographystyle{ieee_fullname}
\bibliography{egbib}
}

\end{document}